\documentclass[10pt,twocolumn,letterpaper]{article}

\usepackage{cvpr}
\usepackage{times}
\usepackage{epsfig}
\usepackage{graphicx}
\usepackage{amsmath}
\usepackage{amssymb}

\usepackage{graphicx}
\graphicspath{{./images/}}
\DeclareGraphicsExtensions{.pdf,.jpeg,.jpg,.png}

\usepackage{float}

\usepackage{subcaption}
\usepackage[font=small,labelfont=bf]{caption}

\usepackage{algorithm}
\usepackage{algpseudocode}
\usepackage{array}

\usepackage{framed}

\usepackage[normalem]{ulem}
\usepackage{amsmath, amsthm, amssymb}
\usepackage{mathtools}
\usepackage{gensymb} % degree symbol

\newcommand{\KL}[1]{\textbf{KL}}
\newcommand{\dist}[1]{{#1}'}
\newcommand{\stab}{\textrm{stability}}
\newcommand{\gty}{\hat{\tb{y}}}

\newcommand{\thumb}{\textsc{thumb}}
\newcommand{\crop}{\textsc{crop}}
\newcommand{\jpeg}{\textsc{jpeg}}
\newcommand{\JPEG}{\jpeg}

\newcommand{\ra}{\textbf{$\rightarrow$}}

\newcommand{\tempnewpage}{}

\let\originalleft\left
\let\originalright\right
\renewcommand{\left}{\mathopen{}\mathclose\bgroup\originalleft}
\renewcommand{\right}{\aftergroup\egroup\originalright}

\newcommand{\refn}[1]{(\ref{#1})}

\newcommand{\brck}[1]{\ensuremath{\left(#1\right)}}

\newcommand{\brckcur}[1]{\ensuremath{\left\{#1\right\}}}

\newcommand{\norm}[1]{\ensuremath{||#1||}}

\newcommand{\be}{\begin{equation}}
\newcommand{\ee}{\end{equation}}
\newcommand{\bali}{\begin{eqnarray*}}
\newcommand{\eali}{\end{eqnarray*}}
\newcommand{\eq}[1]{\begin{align}#1\end{align}}

\newcommand{\iitem}[1]{\begin{itemize}#1\end{itemize}}

\newcommand{\calD}{\mathcal{D}}

\newcommand{\calI}{\mathcal{I}}

\newcommand{\calN}{\mathcal{N}}

\newcommand{\trm}[1]{\textnormal{#1}}

\newcommand{\tb}[1]{\textbf{#1}}
\newcommand{\ti}[1]{\textit{#1}}

\usepackage{color}
\definecolor{blue}{rgb}{0,0,.7}
\definecolor{red}{rgb}{.7,0,0}
\definecolor{orange}{rgb}{1,.6,0}
\definecolor{purple}{rgb}{.4,0,.5}
\definecolor{brown}{rgb}{.4,.2,.1}
\definecolor{green}{rgb}{0,.5,0}

\newcommand{\Red}[1]{{\color{red}#1}}
\newcommand{\Green}[1]{{\color{green}#1}}

\usepackage[pagebackref=true,breaklinks=true,letterpaper=true,colorlinks,bookmarks=false]{hyperref}

\cvprfinalcopy % *** Uncomment this line for the final submission

\def\cvprPaperID{1851} % *** Enter the CVPR Paper ID here

\ifcvprfinal\fi

\begin{document}

%%%%%%%%% TITLE
\title{Improving the Robustness of Deep Neural Networks via Stability Training}

\author{Stephan Zheng \\
Google, Caltech\\
{\tt\small stzheng@caltech.edu}
\and
Yang Song\\
Google\\
{\tt\small yangsong@google.com}
\and
Thomas Leung\\
Google\\
{\tt\small leungt@google.com}
\and
Ian Goodfellow\\
Google\\
{\tt\small goodfellow@google.com}
}

\maketitle

\begin{abstract}

In this paper we address the issue of output instability of deep neural networks: small perturbations in the visual input can significantly distort the feature embeddings and output of a neural network. Such instability affects many deep architectures with state-of-the-art performance on a wide range of computer vision tasks.
We present a general stability training method to stabilize deep networks against small input distortions that result from various types of common image processing, such as compression, rescaling, and cropping.
We validate our method by stabilizing the state-of-the-art Inception architecture \cite{szegedy_going_2015} against these types of distortions.
In addition, we demonstrate that our stabilized model gives robust state-of-the-art performance on large-scale near-duplicate detection, similar-image ranking, and classification on noisy datasets.
\end{abstract}
\section{Introduction}

\begin{figure}[!ht]
\centering
\includegraphics[width=0.45\textwidth]{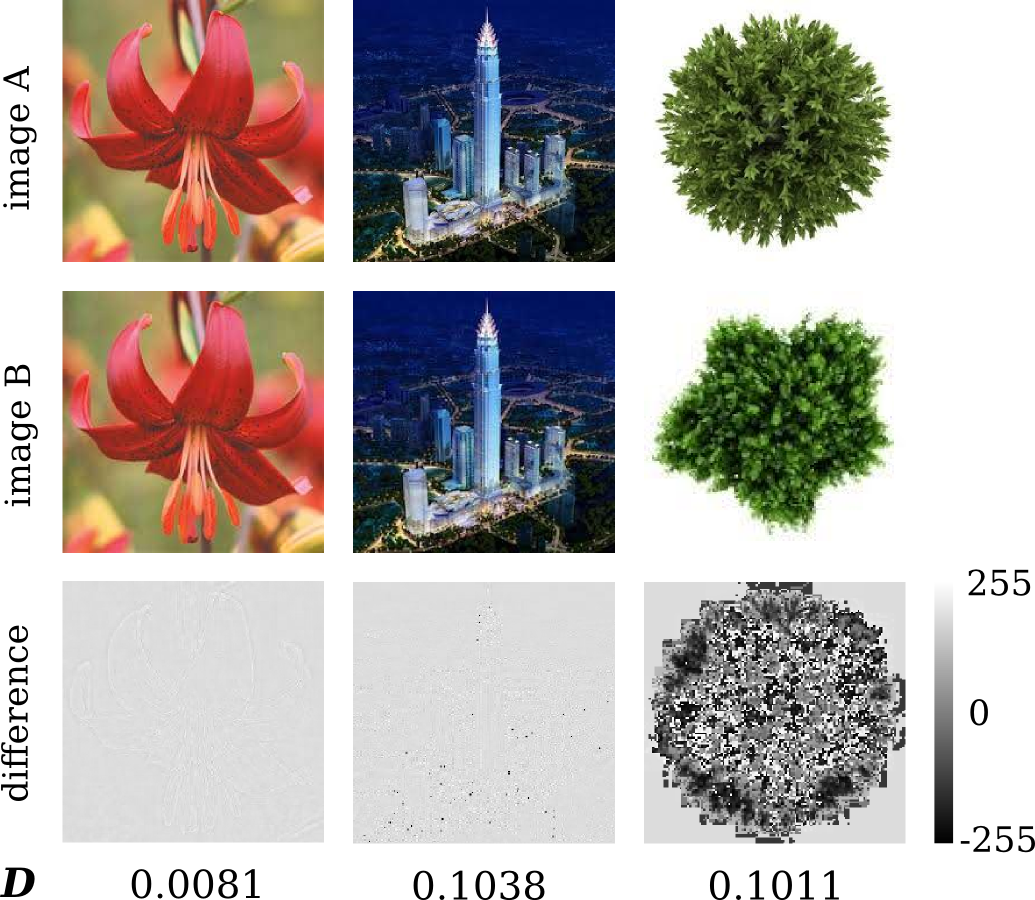}
\caption{Near-duplicate images can confuse state-of-the-art neural networks due to feature embedding instability.
Left and middle columns: near-duplicates with small (left) and large (middle) feature distance. Image A is the original, image B is a JPEG version at quality factor 50. Right column: a pair of dissimilar images.
In each column we display the pixel-wise difference of image A and image B, and the feature distance $D$ \cite{wang_learning_2014}.
Because the feature distances of the middle near-duplicate pair and the dissimilar image pair are comparable, near-duplicate detection using a threshold on the feature distance will confuse the two pairs.
%
%\CORR{I would delete the following sentence.}
%If the detection threshold is too low, the middle pair is not classified as near-duplicate, or the right pair is mistaken as near-duplicate if the threshold is too high.
%
}
\label{fig:fp-example-jpeg}
\end{figure}

Deep neural networks learn feature embeddings of the input data that enable state-of-the-art performance in a wide range of computer vision tasks, such as visual recognition \cite{krizhevsky_imagenet_2012, szegedy_going_2015} and similar-image ranking \cite{wang_learning_2014}.
Due to this success, neural networks are now routinely applied to vision tasks on large-scale \ti{un-curated} visual datasets that, for instance, can be obtained from the Internet. Such un-curated visual datasets often contain small distortions that are undetectable to the human eye, due to the large diversity in formats, compression, and manual post-processing that are commonly applied to visual data in the wild. These lossy image processes do not change the correct ground truth labels and semantic content of the visual data, but can significantly confuse feature extractors, including deep neural networks. Namely, when presented with a pair of indistinguishable images, state-of-the-art feature extractors can produce two significantly different outputs.

In fact, current feature embeddings and class labels are not robust to a large class of small perturbations. Recently, it has become known that intentionally engineered imperceptible perturbations of the input can change the class label output by the model \cite{goodfellow_explaining_2014,szegedy_intriguing_2013}. A scientific contribution of this paper is the demonstration that these imperceptible perturbations can also occur without being contrived and widely occur due to compression, resizing, and cropping corruptions in visual input.

As such, output instability poses a significant challenge for the large-scale application of neural networks because high performance at large scale requires robust performance on noisy visual inputs.
Feature instability complicates tasks such as near-duplicate detection, which is essential for large-scale image retrieval and other applications. In near-duplicate detection, the goal is to detect whether two given images are visually similar or not. When neural networks are applied to this task, there are many failure cases due to output instability. For instance, Figure \ref{fig:fp-example-jpeg} shows a case where a state-of-the-art deep network cannot distinguish a pair of near-duplicates \cite{wang_learning_2014} and a pair of dissimilar images.

% Output instability affects many large-scale visual understanding tasks. In recent years, deep neural networks have achieved state-of-the-art performance on a wide range of visual tasks, such as image classification on curated visual data. However, deep networks still perform inconsistently when used on noisy un-curated visual data.

\begin{figure}[t]
\centering
\includegraphics[width=240pt]{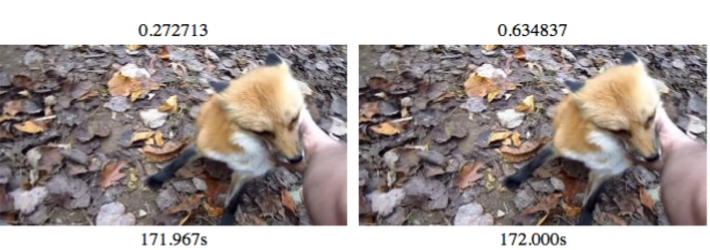}
\caption{Visually similar video frames can confuse state-of-the-art classifiers: two neighboring frames are visually indistinguishable, but can lead to very different class predictions.
The class score for 'fox' is significantly different for the left frame ($27\%$) and right frame ($63\%$), which causes only the fox in the right image to be correctly recognized, using any reasonable confidence threshold (e.g. $> 50\%$).}
\label{fig:fp-class-instability}
\end{figure}

Analogously, class label instability introduces many failure cases in large-scale classification and annotation. For example, unstable classifiers can classify neighboring video-frames inconsistently, as shown in Figure \ref{fig:fp-class-instability}. In this setting, output instability can cause large changes in label scores of a state-of-the-art convolutional neural network on consecutive video-frames that are indistinguishable.

% Another class of natural perturbations arises when producing thumbnail versions or cropping an image by a few pixels. As the visual field of the neural network input layer is fixed, these smaller versions of the original image need to be up-scaled or padded when shown to the neural network. Such up-scaling and interpolation methods introduce small changes compared to the original image that lead to inconsistent model output. An example of resizing artifacts is shown in Figure \ref{fig:fp-example-resize}.

The goal of this paper is to propose a general approach to stabilize machine learning models, in particular deep neural networks, and make them more robust to visual perturbations. To this end, we introduce a fast and effective \ti{stability training} technique that makes the output of neural networks significantly more robust, \ti{while maintaining or improving state-of-the-art performance on the original task}.
To do so, our method operates through two mechanisms: 1) introducing an additional stability training objective and 2) training on a large class of distorted copies of the input.
The goal of this approach is to force the prediction function of the model to be more constant around the input data, while preventing underfitting on the original learning objective.
In summary, our contributions are as follows:
\iitem{
\item
We propose stability training as a general technique that improves model output stability while maintaining or improving the original performance. Our method is fast in practice and can be used at a minimal additional computational cost.
\item We validate our method by stabilizing state-of-the-art classification and ranking networks based on the Inception architecture \cite{szegedy_going_2015, wang_learning_2014}. We evaluate on three tasks: near-duplicate image detection, similar-image ranking, and image classification.
\item We show the impact of stability training by visualizing what perturbations the model has become robust to.
\item Finally, we show that stabilized networks offer robust performance and significantly outperform unstabilized models on noisy and corrupted data.
}
%
% The rest of this paper is structured as follows. In section \ref{sec:relatedwork} we discuss related work.
% We describe our stabilization method in section \ref{sec:method} and our experimental setup in section \ref{sec:experiments}. Finally, qualitative and quantitative results are presented in section \ref{sec:results}.

\section{Related work}\label{sec:relatedwork}

\tb{Adversarial examples.} Recently, several machine learning
algorithms were found to have extreme instability against
\ti{contrived} input perturbations \cite{szegedy_intriguing_2013}
called adversarial examples.
% Note: Nguyen_2015_CVPR isn't about instability to perturbation,
% it's about how the model responds to complete garbage inputs
% that don't resemble the data at all. So it's not very relevant
% here.
An open question remained as to whether such small perturbations
that change the class label could occur without intentional
human intervention. In this work, we document that they do in
fact occur.
Previous work has shown that training a classifier to resist
adversarial perturbation can improve its performance on both the original data and on perturbed data \cite{goodfellow_explaining_2014,miyato_distributional_2015}.
We extend this approach by training our feature embeddings to
resist the naturally occurring perturbations that are far more
common in practice.

Furthermore, our work differs drastically from \cite{Nguyen_2015_CVPR}, which is about how a model responds to intentionally contrived inputs that don't resemble the original data at all. In contrast, in this paper we consider the stability to practically widely occurring perturbations.

\tb{Data augmentation.} A natural strategy to improve label stability is to augment the training data with \ti{hard positives}, which are examples that the prediction model does not classify correctly with high confidence, but that are visually similar to easy positives. Finding such hard positives in video data for data augmentation has been used in \cite{MisraExemplarSelection, kuznetsova_expanding_2015, prest_learning_2012} and has been found to improve predictive performance and consistency. As such, data augmentation with hard positives can confer output stability on the classes of perturbations that the hard positives represent. However, our work differs from data augmentation in two ways.
Firstly, we take a general approach by proposing a method that intends to make model performance more robust to various types of natural perturbations.
Secondly, our proposed method does not use the extra generated samples as training examples for the original prediction task, but only for the stability objective.

% \section{Model instability and ineffectiveness of natural data augmentation}\label{sec:problem}

 \tempnewpage
\section{Stability training}\label{sec:method}

We now present our stability training approach, and how it can be applied to learn robust feature embeddings and class label predictions.

\subsection{Stability objective}

Our goal is to stabilize the output $f(x)\in\mathbb{R}^m$ of a neural network $\calN$ against small natural perturbations to a natural image $x\in [0,1]^{w\times h}$ of size $w\times h$, where we normalize all pixel values. Intuitively, this means that we want to formulate a training objective that flattens $f$ in a small neighborhood of any natural image $x$: if a perturbed copy $x'$ is close to $x$, we want $f(x)$ to be close to $f(x')$, that is
\eq{\label{eq:stability_condition} \forall \trm{} x': d(x, x') \trm{ small}   \Leftrightarrow D\brck{f(x), f(x')} \trm{ small}.}
Here $d$ is the distance on $[0,1]^{w\times h}$ and $D$ is an appropriate distance measure in feature space.

% $\delta$ and $\epsilon$ are small parameters that formally define the range of natural image with perturbations that we consider,

Given a training objective $L_0$ for the original task (e.g. classification, ranking), a reference input $x$ and a perturbed copy $x'$, we can implement the stability objective \refn{eq:stability_condition} as:
\eq{
L(x,x'; \theta) &= L_{0}(x; \theta) + \alpha L_{\stab}(x,x'; \theta), \label{eq:full_loss}\\
L_{\stab}(x, x'; \theta) &= D\brck{f(x), f(x')}, \label{eq:stability_loss}
}
where $\alpha$ controls the strength of the stability term and $\theta$ denotes the weights of the model $\calN$. The stability objective $L_{\stab}$ forces the output $f(x)$ of the model to be similar between the original $x$ and the distorted copy $x'$. Note that \ti{our approach differs from data augmentation}: we do not evaluate the original loss $L$ on the distorted inputs $x'$. This is required to achieve both output stability and performance on the original task, as we explain in \ref{sec:data-aug}.

Given a training dataset $\calD$, stability training now proceeds by finding the optimal weights $\theta^*$ for the training objective \refn{eq:full_loss}, that is, we solve
\eq{\theta^*=\underset{\theta}{\trm{argmin}} \sum_{x_i\in \calD, d(x_i,x_i')<\epsilon}L(x_i,x'_i; \theta).}
To fully specify the optimization problem, we firstly need a mechanism to generate, for each training step, for each training sample $x_i$, a random perturbed copy $x'_i$. Secondly, we need to define the distance $D$, which is task-specific.

% \INST{Discuss issues: how big are $\delta, \epsilon$? Look at the JPEG versions: what are delta and espilon there? What directions does it extend into? The boundary is not precise: you can have a big perturbation of a single pixel, and still want the same answer. Perturbations are either localized, or small and dispersed. }

\subsection{Sampling perturbed images $x'$}\label{sec:data-aug}

\tb{Sampling using Gaussian noise.} During training, at every training step we need to generate perturbed versions $x'$ of a clean image $x$ to evaluate the stability objective \refn{eq:stability_loss}.

A natural approach would be to augment the training data with examples with explicitly chosen classes of perturbation that the model should be robust against. However, it is hard to obtain general robustness in this way, as there are many classes of perturbations that cause output instability, and model robustness to one class of perturbations does not confer robustness to other classes of perturbations.

Therefore, we take a general approach and use a sampling mechanism that adds pixel-wise uncorrelated Gaussian noise $\epsilon$ to the visual input $x$. If $k$ indexes the raw pixels, a new sample is given by:
\eq{%
\dist{x}_k = x_k+\epsilon_k, \hspace{5pt} \epsilon_k \sim \calN\brck{0, \sigma_k^2},  \hspace{5pt} \sigma_k>0,\label{eq:gaussian-noise}
}
where $\sigma_k^2$ is the variance of the Gaussian noise at pixel $k$. In this work, we use uniform sampling $\sigma_k = \sigma$ to produce unbiased samples of the neighborhood of $x$, using the variance $\sigma^2$ as a hyper-parameter to be optimized.

\begin{figure}[!t]
\centering
\includegraphics[width=116pt]{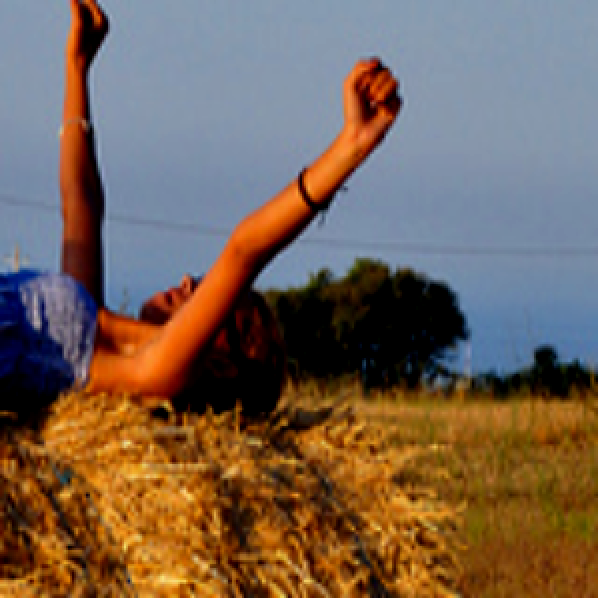}
\includegraphics[width=116pt]{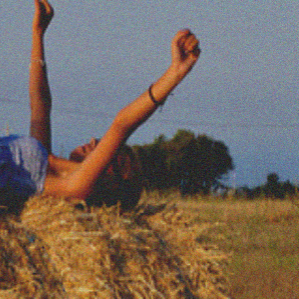}
\caption{Examples of reference and distorted training images used for stability training. Left: an original image $x$. Right: a copy $x'$ perturbed with pixel-wise uncorrelated Gaussian noise with $\sigma=0.06$, in normalized pixel values. During stability training, we use dynamically sampled copies $x'$ together with the stability loss \refn{eq:stability_loss} to flatten the prediction function $f$ around the original image $x$.
}
\label{fig:new-neardupes}
\end{figure}

\tb{Preventing underfitting.}
Augmenting the training data by adding uncorrelated Gaussian noise can potentially simulate many types of perturbations. Training on these extra samples could in principle lead to output robustness to many classes of perturbations.
%
%However, the number of perturbation classes that is sampled from grows exponentially with the size of the input\footnote{This contrasts with data augmentation with large or global correlated perturbations, such as translations, mirroring and large rotations, which only use a sub-exponential number of perturbation classes.}
%.
%Therefore, sampling with Gaussian noise is much more prone to underfitting and can %quickly lead to decreased performance. Intuitively, data augmentation with Gaussian noise %over-regularizes the model such that the model cannot determine anymore what an %unperturbed natural image looks like.
However, we found that training on a dataset augmented by Gaussian perturbation
leads to underfitting, as shown in Table \ref{table:underfitting}.
%
% Similar underfitting has been documented in the music domain \cite{grill_music_boundary_2015}.
%
%, perhaps because so many of the Gaussian perturbations are unrelated to the variations that occur in the test set.
%
\begin{table}[t]
    \centering
    \begin{tabular}{c|ccc}
    Gaussian noise strength $\sigma$ & 0.0 & 0.1 & 0.2 \\
    \hline
    Triplet ranking score @ top-30 & 7,312 & 6,300 & 5,065
    \end{tabular}
    \caption{Underfitting by data augmentation with Gaussian noise on an image ranking task (higher score is better), see section \ref{sec:tripletranking} for details. The entry with $\sigma=0.0$ is the model without data augmentation.}
    \label{table:underfitting}
\end{table}
To prevent such underfitting, we do \ti{not} evaluate the original loss $L_0$ on the perturbed images $x'$ in the full training objective \refn{eq:full_loss}, but only evaluate the stability loss \refn{eq:stability_loss} on both $x$ and $x'$. This approach differs from data augmentation, where one would evaluate $L_0$ on the extra training samples as well. It enables achieving both output stability and maintaining high performance on the original task, as we validate empirically.

% Secondly, as stability training requires twice as many inference steps due to the extra samples $x'$, it is costly to train on \refn{eq:full_loss} from the beginning of the training procedure. Instead, it is more convenient and equally effective to turn on the stability term \refn{eq:stability_loss} when training has converged on the original objective $L_0$.
%
% \CORR{However, in that case, evaluating the original loss $L_0$ on $x'$ as well does not lead to improvement on $L_0$, but rather leads to noisy gradients that generically move the model away from the optimal values $\theta^*$ and re-introduce underfitting. To maintain optimal performance, we therefore evaluate the original loss $L_0$ only on the original samples $x_i$.}

\subsection{Stability for feature embeddings}\label{sec:stab-feat-emb}

We now show how stability training can be used to obtain stable feature embeddings. In this work, we aim to learn feature embeddings for robust similar-image detection. To this end, we apply stability training in a ranking setting. The objective for similar-image ranking is to learn a feature representation $f(x)$ that detects visual image similarity \cite{wang_learning_2014}. This learning problem is modeled by considering a \ti{ranking triplet} of images $(q,p,n)$: a \ti{query} image $q$, a \ti{positive} image $p$ that is visually similar to $q$, and a \ti{negative} image $n$ that is less similar to $q$ than $p$ is.

%Using an abstract similarity measure $r$ on images, we have
%
%\eq{r(q,p) < r(q,n).\label{eq:image-ranking}}
%
The objective is to learn a feature representation $f$ that respects the triplet ranking relationship in feature space, that is,
\eq{D(f(q), f(p)) + g < D(f(q), f(n)),\hspace{5pt} g>0, \label{eq:feature-ranking}}
where $g$ is a margin and $D$ is the distance. We can learn a model for this objective by using a hinge loss:
\eq{
&L_{0}(q,p,n) = \notag\\
&\max\brck{0, g + D(f(q),f(p)) - D(f(q),f(n))}.\label{eq:hingeloss}
}
In this setting, a natural choice for the similarity metric $D$ is the $L_2$-distance. The stability loss is,
\eq{
L_{\stab}\brck{x,x'}=\norm{f(x)-f(x')}_2.
}

To make the feature representation $f$ stable using our approach, we sample
triplet images $\brck{q',p',n'}$ close to the reference $\brck{q,p,n}$, by applying \refn{eq:gaussian-noise} to each image in the triplet.
% \eq{q'=q+\,\hspace{5pt} p'=p + \epsilon,\hspace{5pt} n' = n.}
% We do so by only sampling a perturbed copy of the positive image $p$, that is

\subsection{Stability for classification}

We also apply stability training in the classification setting to learn stable prediction labels for visual recognition. For this task, we model the likelihood $P\brck{\tb{y}|x;\theta}$ for a labeled dataset $\brckcur{(x_i, \gty_i)}_{i\in\calI}$, where $\gty$ represents a vector of ground truth binary class labels and $i$ indexes the dataset. The training objective is then to minimize the standard cross-entropy loss
\eq{%
L_0(x;\theta) = -\sum_{j} \hat{y}_j\log P\brck{y_j | x; \theta},%
}
where the index $j$ runs over classes.
To apply stability training, we use the KL-divergence as the distance function $D$:
\eq{%
L_{\stab}(x, x'; \theta) %\trm{KL}(P, P') \notag\\
= -\sum_j P\brck{y_j|x;\theta} \log P\brck{y_j|x';\theta},
}
which measures the correspondence between the likelihood on the natural and perturbed inputs.

\textbf{} \tempnewpage
\section{Implementation}\label{sec:experiments}

\subsection{Network}\label{sec:network}

\tb{Base network.} In our experiments, we use a state-of-the-art convolutional neural network architecture, the Inception network \cite{szegedy_going_2015} as our base architecture. Inception is formed by a deep stack of composite layers, where each composite layer output is a concatenation of outputs of convolutional and pooling layers. This network is used for the classification task and as a main component in the triplet ranking network.

\tb{Triplet ranking network.} Triplet ranking loss \refn{eq:hingeloss} is used train feature embeddings for image similarity and for near duplicate image detection, similar to \cite{wang_learning_2014}.
This network architecture uses an Inception module (while in \cite{wang_learning_2014}, a network like \cite{krizhevsky_imagenet_2012} is used) to process every input image $x$ at full resolution and uses 2 additional low-resolution towers. The outputs of these towers map into a 64-dimensional $L_2$-normalized embedding feature $f(x)$. These features are used for the ranking task: for each triplet of images $\brck{q,p,n}$, we use the features $\brck{f(q), f(p), f(n)}$ to compute the ranking loss and train the entire architecture.

\begin{figure}[!t]
\centering
\includegraphics[width=230pt]{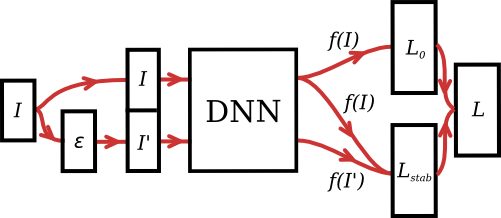}
\caption{The architecture used to apply stability training to any given deep neural network. The arrows display the flow of information during the forward pass. For each input image $I$, a copy $I'$ is perturbed with pixel-wise independent Gaussian noise $\epsilon$. Both the original and perturbed version are then processed by the neural network. The task objective $L_0$ is only evaluated on the output $f(I)$ of the original image, while the stability loss $L_{\stab}$ uses the outputs of both versions. The gradients from both $L_0$ and $L_{\stab}$ are then combined into the final loss $L$ and propagated back through the network. For triplet ranking training, three images are processed to compute the triplet ranking objective.}
\label{fig:stability_training_network}
\end{figure}

\tb{Stability training.} It is straightforward to implement stability training for any given neural network by adding a Gaussian perturbation sampler to generate perturbed copies of the input image $x$ and an additional stability objective layer.
This setup is depicted in Figure \ref{fig:stability_training_network}.

\subsection{Distortion types}\label{sec:evaldistortions}

\begin{figure*}[!ht]
\begin{center}
\begin{subfigure}[b]{\textwidth}\label{ref_label2}\centering
\includegraphics[width=0.115\textwidth]{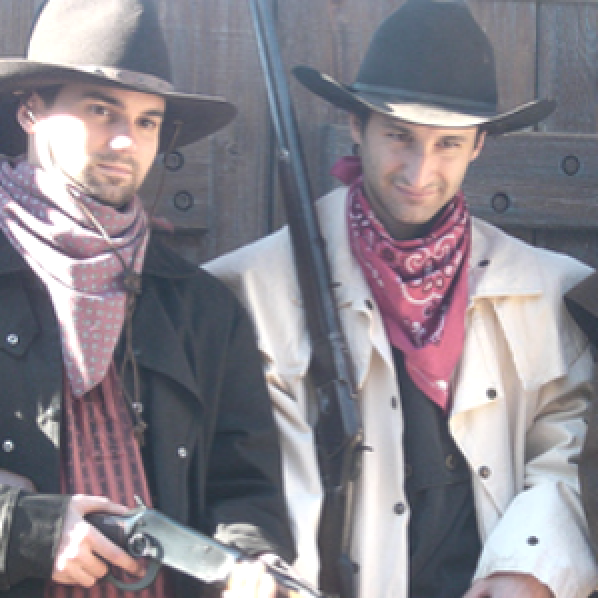}
\includegraphics[width=0.115\textwidth]{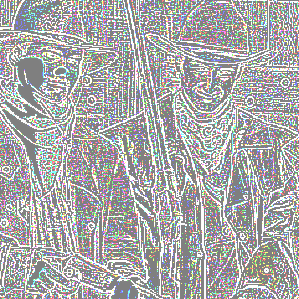}
\includegraphics[width=0.115\textwidth]{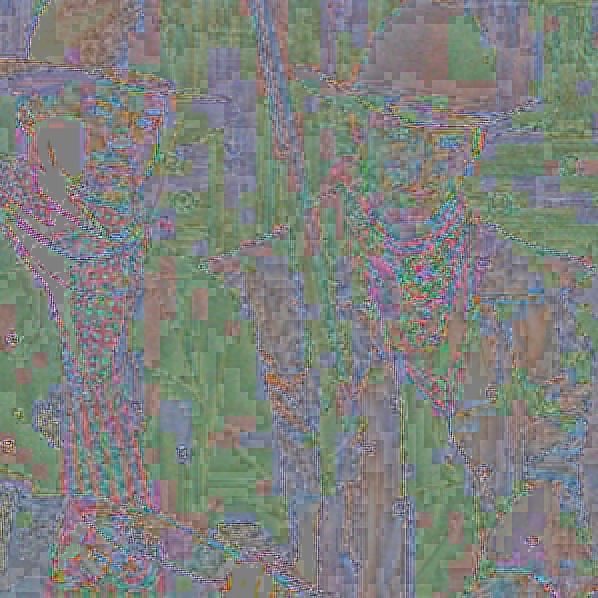}
\includegraphics[width=0.115\textwidth]{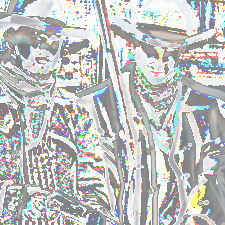}\hspace{19pt}
\includegraphics[width=0.115\textwidth]{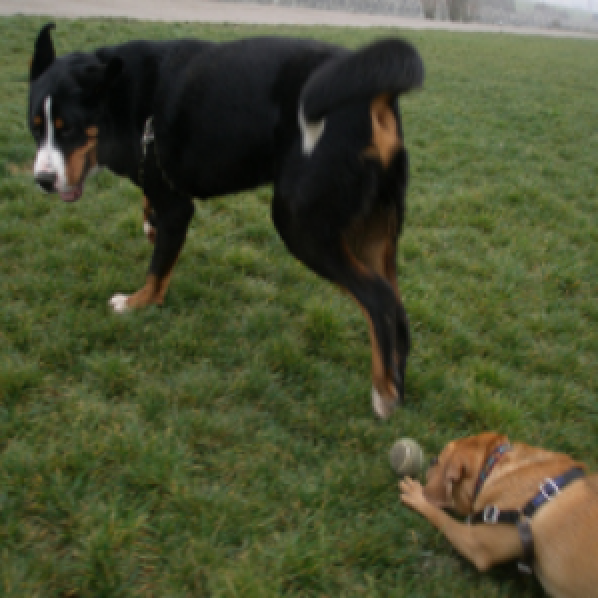}
\includegraphics[width=0.115\textwidth]{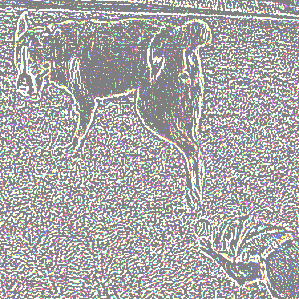}
\includegraphics[width=0.115\textwidth]{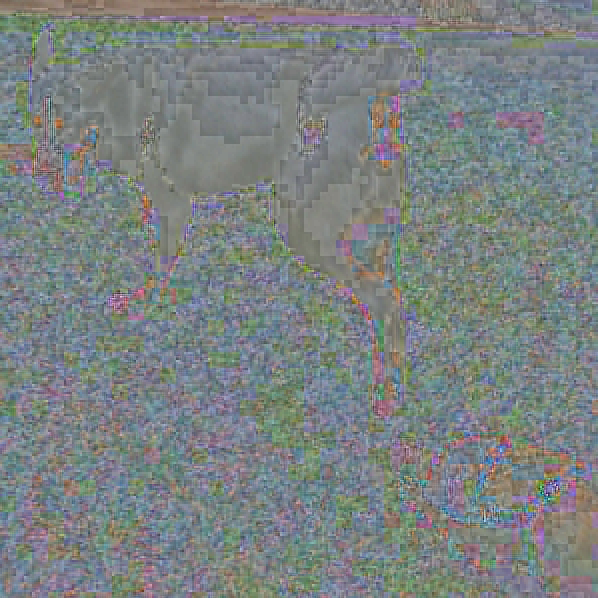}
\includegraphics[width=0.115\textwidth]{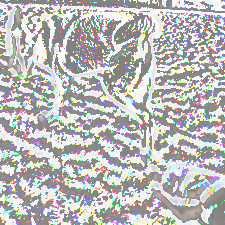}\\
\includegraphics[width=0.115\textwidth]{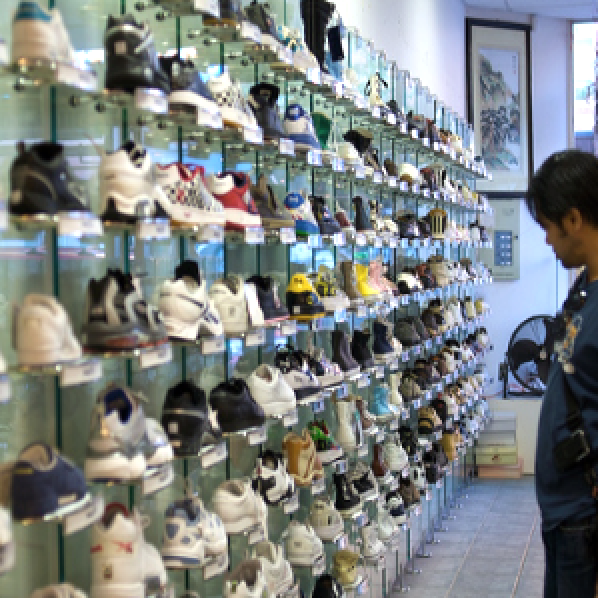}
\includegraphics[width=0.115\textwidth]{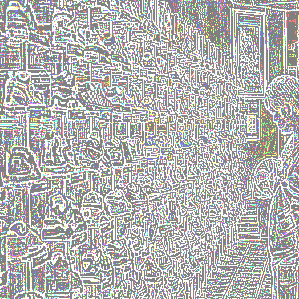}
\includegraphics[width=0.115\textwidth]{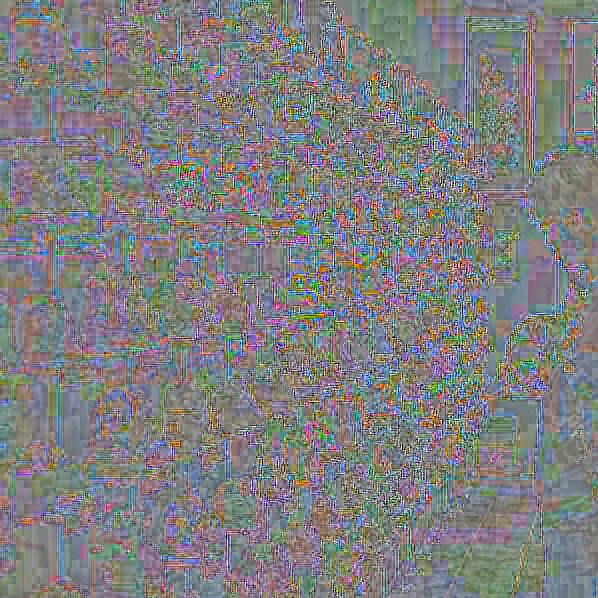}
\includegraphics[width=0.115\textwidth]{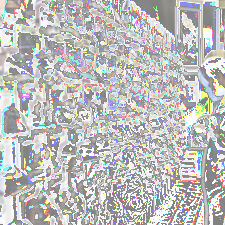}\hspace{19pt}
\includegraphics[width=0.115\textwidth]{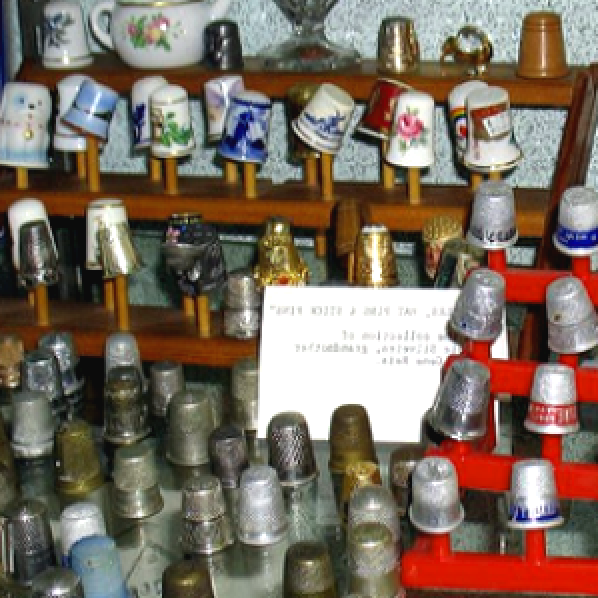}
\includegraphics[width=0.115\textwidth]{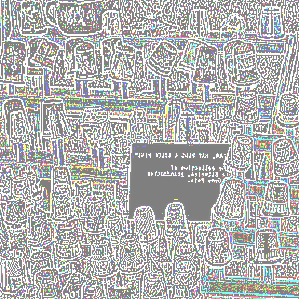}
\includegraphics[width=0.115\textwidth]{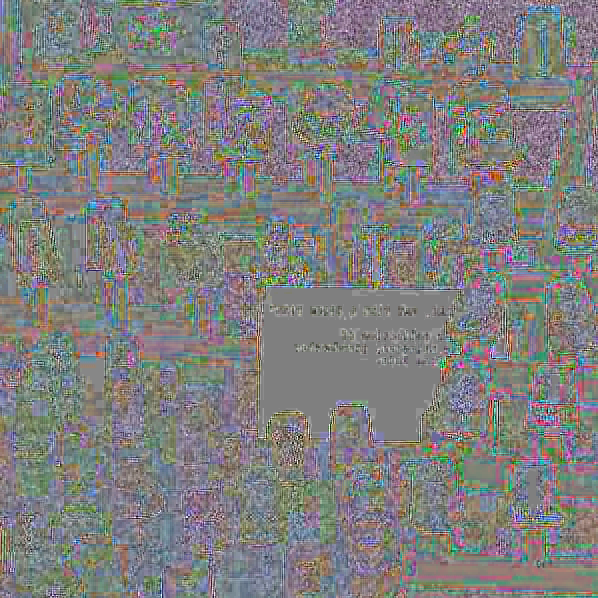}
\includegraphics[width=0.115\textwidth]{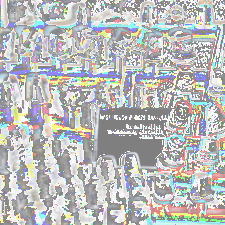}\\
\includegraphics[width=0.115\textwidth]{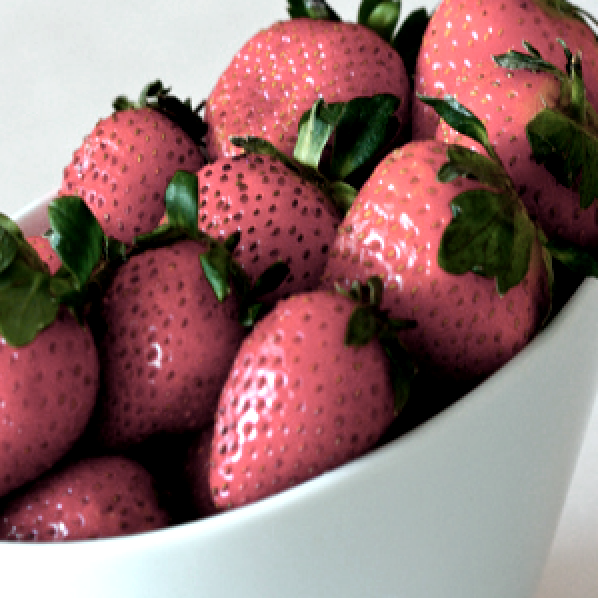}
\includegraphics[width=0.115\textwidth]{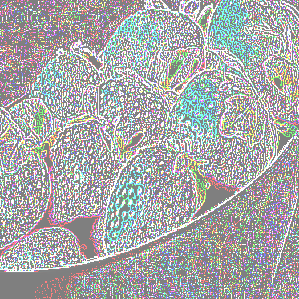}
\includegraphics[width=0.115\textwidth]{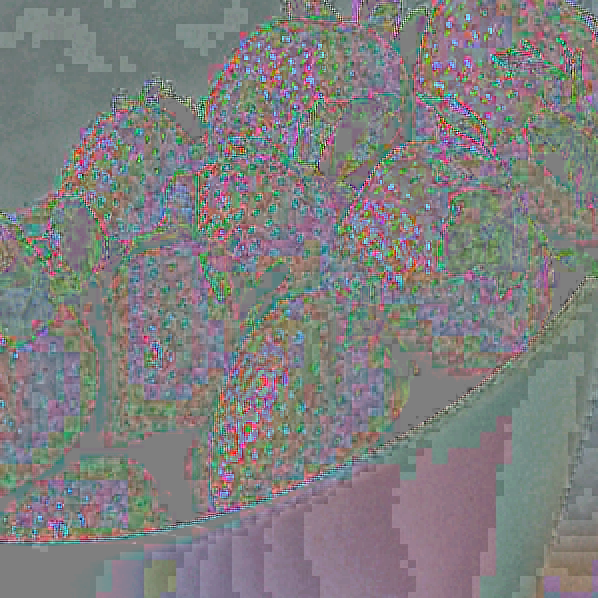}
\includegraphics[width=0.115\textwidth]{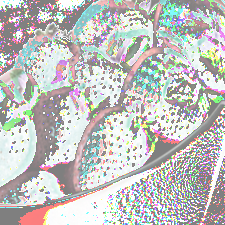}\hspace{19pt}
\includegraphics[width=0.115\textwidth]{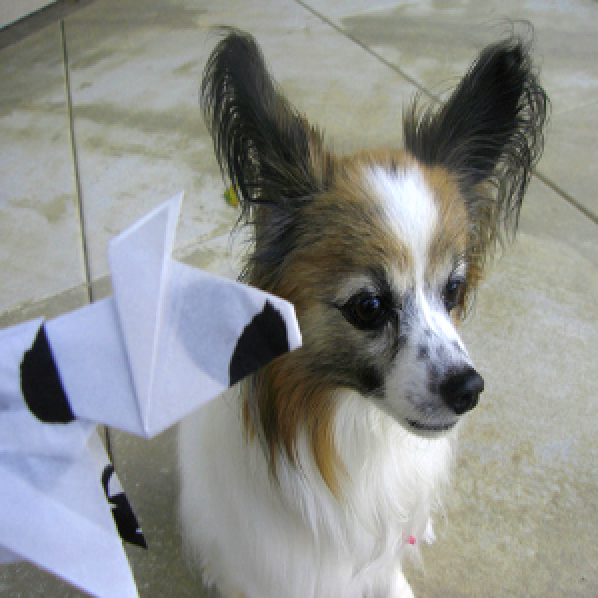}
\includegraphics[width=0.115\textwidth]{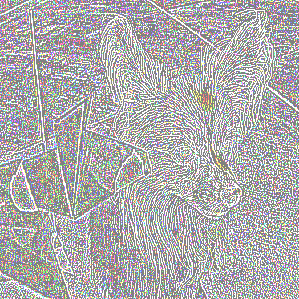}
\includegraphics[width=0.115\textwidth]{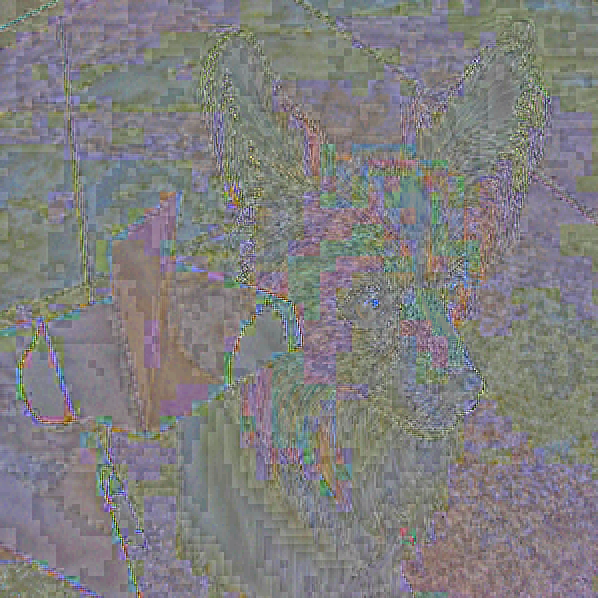}
\includegraphics[width=0.115\textwidth]{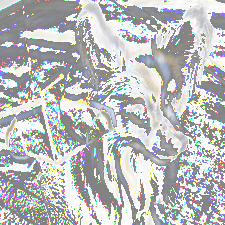}
\end{subfigure}
\end{center}
%   \subfloat[Caption 1]{\label{ref_label1}\includegraphics[width=0.5\textwidth]{path/to/figure1}}
\caption{
Examples of natural distortions that are introduced by common types of image processing. From left to right: original image (column 1 and 5), pixel-wise differences from the original after different forms of transformation: thumbnail downscaling to $225\times 225$ (column 2 and 6), \jpeg{} compression at quality level 50\% (column 3 and 7) and random cropping with offset $10$ (column 4 and 8). For clarity, the \jpeg{} distortions have been up-scaled by $5\times$.
Random cropping and thumbnail resizing introduce distortions that are structured and resemble the edge structure of the original image. In contrast, \jpeg{} compression introduces more unstructured noise.
%
% We evaluate the performance of a network with stability training on these types of natural distortions.
}
\label{fig:eval_distortions}
\end{figure*}

To demonstrate the robustness of our models after stability training is deployed, we evaluate the ranking, near-duplicate detection and classification performance of our stabilized models on both the original and transformed copies of the evaluation datasets. To generate the transformed copies, we apply visual perturbations that widely occur in real-world visual data and that are a result of lossy image processes.

\tb{\JPEG{} compression.} \JPEG{} compression is a commonly used lossy compression method that introduces small artifacts in the image. The extent and intensity of these artifacts can be controlled by specifying a quality level $q$. In this work, we refer to this as \JPEG-$q$.

\tb{Thumbnail resizing.} Thumbnails are smaller versions of a reference image and obtained by downscaling the original image. Because convolutional neural networks use a fixed input size, both the original image and its thumbnail have to be rescaled to fit the input window. Downscaling and rescaling introduces small differences between the original and thumbnail versions of the network input. In this work we refer to this process as \thumb-$A$, where we downscale to a thumbnail with $A$ pixels, preserving the aspect ratio.

\tb{Random cropping.} We also evaluated the performance on perturbations coming from random crops of the original image. This means that we take large crops with window size $w'\times h'$ of the original image of size $w\times h$, using an offset $o>0$ to define $w' = w - o, h'=h-o$. The crops are centered at random positions, with the constraint that the cropping window does not exceed the image boundaries. Due to the fixed network input size, resizing the cropped image and the original image to the input window introduces small perturbations in the visual input, analogous to thumbnail noise. We refer to this process as \crop-$o$, for crops with a window defined by offset $o$.

\subsection{Optimization}
To perform stability training, we solved the optimization problem \refn{eq:full_loss} by training the network using mini-batch stochastic gradient descent with momentum, dropout \cite{srivastava_dropout:_2014}, RMSprop and batch normalization \cite{ioffe_batch_2015}. To tune the hyper-parameters, we used a grid search, where the search ranges are displayed in Table \ref{table:hyperparameters}.

\begin{table}[!ht]\centering
\begin{tabular}{l|c|c}
Hyper-parameter & Start range & End range \\
\hline
Noise standard deviation $\sigma$ & 0.01 & 0.4 \\
Regularization coefficient $\alpha$ & 0.001 & 1.0 \\
Learning rate $\lambda$ & 0.001 & 0.1 \\
% Learning rate decay & 0.8 & 0.99 \\
% Dropout $p$ & 0.01 & 0.5 \\
% RMSProp decay & 0.5 & 0.99 \\
% Gap parameter $g$ & 0.01 & 0.2
% \hline \\
% Gap parameter $g$ & 0.05
\end{tabular}
\caption{Hyper-parameter search range for the stability training experiments.}
\label{table:hyperparameters}
\end{table}

As stability training requires a distorted version of the original training example, it effectively doubles the training batch-size during the forward-pass, which introduces a significant extra computational cost.
To avoid this overhead, in our experiments we first trained the network on the original objective $L_0(x; \theta)$ only and started stability training with $L(x,x';\theta)$ only in the fine-tuning phase. Additionally, when applying stability training, we only fine-tuned the final fully-connected layers of the network.
Experiments indicate that this approach leads to the same model performance as applying stability training right from the beginning and training the whole network during stability training.

% In this phase, the network has been largely optimized for the original objective $L_0$ and stability training will then further fine-tune the network for stability while preserving the performance on the original task, as it does not evaluate $L_0$ on the distorted copies of the input.
%
 \tempnewpage
\section{Experiments}\label{sec:results}

%Recall that our goal is to use stability training to learn robust feature embeddings and class labels for stable image similarity detection and classification.

Here we present experimental results to validate our stability training method and characterize stabilized models.
\iitem{
\item Firstly, we evaluate stabilized features on near-duplicate detection and similar-image ranking tasks.
\item Secondly, we validate our approach of stabilizing classifiers on the ImageNet classification task.
}

% For all experiments, we use un-stabilized features and class predictions as our baseline quantities.

% For all experiments, we present a qualitative and quantitative performance evaluation on both the original and distorted versions of the evaluation dataset.

We use training data as in \cite{wang_learning_2014} to train the feature embeddings for near-duplicate detection and similar-image ranking. For the classification task, training data from ImageNet are used.

%\tb{Triplet ranking dataset.}
%
%For our experiments with stable features, we used ranking features learned through the deep ranking network (see \ref{sec:network}). To this end, we obtained the triplet dataset that was used in \cite{wang_learning_2014}. The training dataset contains 14 million images with a learned pair-wise relevance measure, that acts as a proxy for the ground-truth similar-image ranking labels.
%Additionally, the triplet ranking dataset includes a set of 14,000 triplet ranking labels obtained from human raters and that was only used during evaluation. A subset of 5,000 triplet labels is publicly available on \url{https://sites.google.com/site/imagesimilaritydata/}.

\subsection{Near-duplicate detection}

% To evaluate stabilized feature embeddings, we applied stability training to the deep ranking network (section \ref{sec:tripletranking}) to learn a stabilized feature $f$.
%
\tb{Detection criterion.} We used our stabilized ranking feature to perform near-duplicate detection. To do so, we define the detection criterion as follows: given an image pair $\brck{a,b}$, we say that
\eq{\label{eq:near-dupe-threshold}
a, b \trm{ are near-duplicates } \Longleftrightarrow \norm{f(a)-f(b)}_2 < T,
}
where $T$ is the near-duplicate detection threshold.

% \subsubsection{Triplet dataset}

% For these images, a pairwise relevance measure $\hat{r}$ was derived from a linear combination of visual and semantic features. This relevance measure $\hat{r}$ serves as a proxy for the abstract relevance measure described in \refn{eq:image-ranking}, and is only used during training.

% We use the relevance measure $\hat{r}$ to dynamically generate triplets during training. Since the number of possible triplets is far too large, we use the biased triplet sampling approach from \cite{wang_learning_2014}, to sample suitable triplets for training. In particular, this sampling policy biases the positive image sampling rate towards more relevant positive images and only samples negative \ti{in-class} images that are less relevant than the positive image by a set margin. Here in-class means that the images in the image pair appear in the search results for the \ti{same} Image Search query.

% Additionally, the dataset includes a set of 14.000 triplet ranking labels obtained from human raters. These triplets labels capture, for a triplet $\brck{q, a, b}$, whether $a$ is the positive image $p$ or the negative image $n$, and similarly for $b$. In this way, they capture the ground truth triplet ranking order \refn{eq:image-ranking} and are only used during evaluation. A subset of these 5.000 triplet labels is publicly available on \url{https://sites.google.com/site/imagesimilaritydata/}.

\tb{Near-duplicate evaluation dataset.}
For our experiments, we generated an image-pair dataset with two parts: one set of pairs of near-duplicate images (true positives) and a set of dissimilar images (true negatives).

We constructed the near-duplicate dataset by collecting 650,000 images from randomly chosen queries on Google Image Search. In this way, we obtained a representative sample of un-curated images. We then combined every image with a copy perturbed with the distortion(s) from section \ref{sec:evaldistortions} to construct near-duplicate pairs.
For the set of dissimilar images, we collected 900,000 random image pairs from the top 30 Image Search results for 900,000 random search queries, where the images in each pair come from the \ti{same} search query.

\subsubsection{Experimental results}

\begin{figure*}[t!]
\centering
\includegraphics[width=170pt]{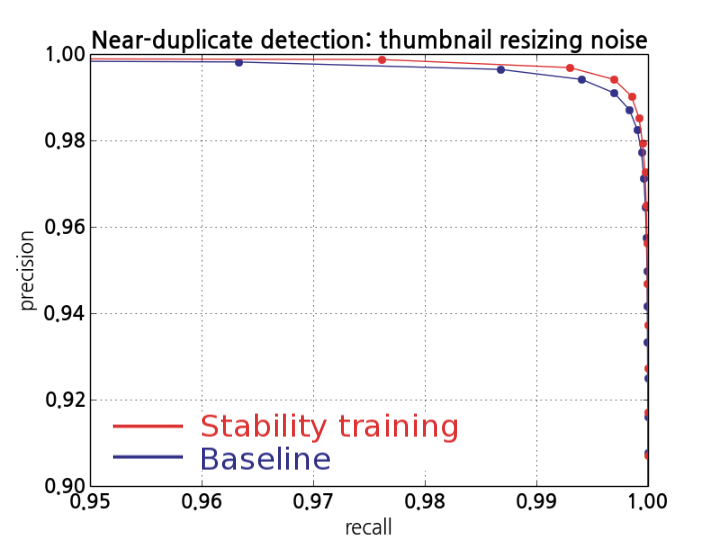}\hspace{-9pt}
\includegraphics[width=170pt]{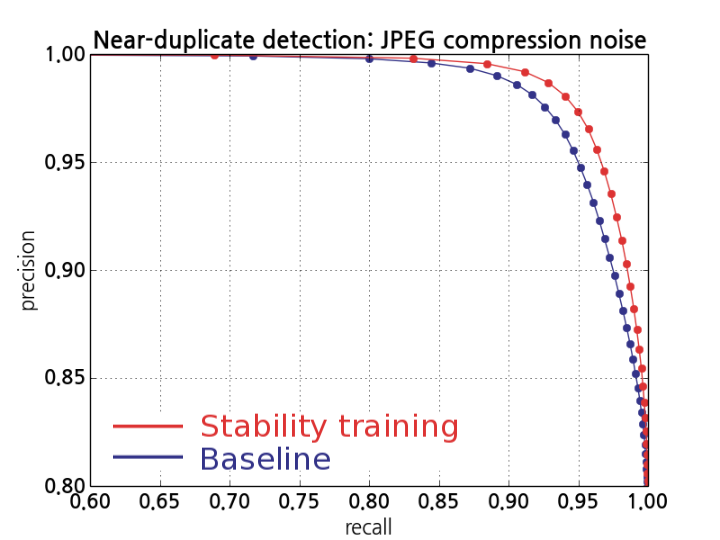}\hspace{-9pt}
\includegraphics[width=170pt]{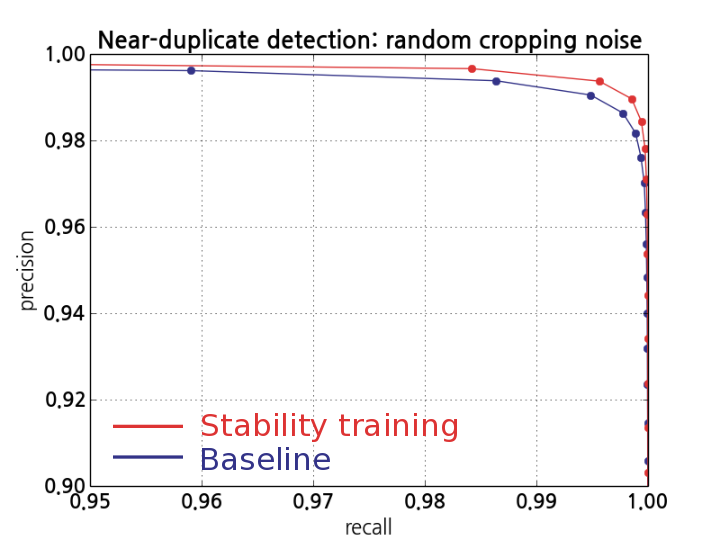}
\caption{
Precision-recall performance for near-duplicate detection using feature distance thresholding on deep ranking features. % on a dataset of 650.000 near-duplicates and 900.000 dissimilar image pairs.
We compare Inception-based deep ranking features (blue), and the same features with stability training applied (red).
Every graph shows the performance using near-duplicates generated through different distortions.
Left: \thumb{}-$50k$. Middle: \jpeg{}-50. Right: \crop{}-10.
% was crop-235 in previous defn of crop
% was \thumb{}-$225$ in prev defn
%
Across the three near-duplicate tasks, the stabilized model significantly improves the near-duplicate detection precision over the baseline model.
}
\label{fig:near-duplicate-pr}
\end{figure*}

\tb{Precision-recall performance.}
To analyze the detection performance of the stabilized features, we report the near-duplicate precision-recall values by varying the detection threshold in \refn{eq:near-dupe-threshold}. Our results are summarized in Figure \ref{fig:near-duplicate-pr}.
The stabilized deep ranking features outperform the baseline features for all three types of distortions, for all levels of fixed recall or fixed precision. Although the baseline features already offer very high performance in both precision and recall on the near-duplicate detection task, the stabilized features significantly improve precision across the board.
For instance, recall increases by 1.0\% at 99.5\% precision for thumbnail near-duplicates,
and increases by 3.0\% at 98\% precision for \jpeg{} near-duplicates.
This improved performance is due to the improved robustness of the stabilized features, which enables them to correctly detect near-duplicate pairs that were confused with dissimilar image pairs by the baseline features, as illustrated in Figure \ref{fig:fp-example-jpeg}.

\begin{figure}[!t]
\centering
\includegraphics[width=0.47\textwidth,height=130pt]{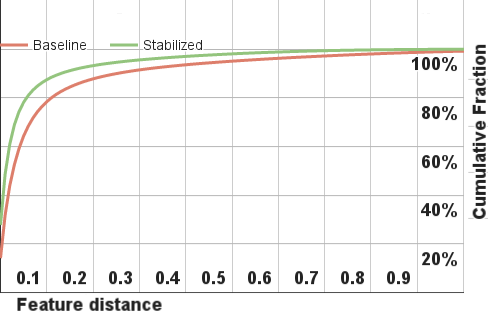}
\caption{
Cumulative distribution of the deep ranking feature distance $D(f(x_i),f(x_i')) = \norm{ f(x_i)-f(x_i') }_2$ for near-duplicate pairs $(x_i,x_i')$.
Red: baseline features,  76\% of distribution $< 0.1$. %
Green: stabilized features using stability training with $\alpha=0.1, \sigma=0.2$, 86\% of distribution $< 0.1$.
The feature distances are computed over a dataset of 650,000 near-duplicate image pairs (reference image and a \JPEG{}-50 version). Applying stability training makes the distribution of $D(f(x), f(x'))$ more concentrated towards 0 and hence makes the feature $f$ significantly more stable.
}
\label{fig:new-neardupes-rank}
\end{figure}

\tb{Feature distance distribution.}
To analyze the robustness of the stabilized features, we show the distribution of the feature distance $D(f(x), f(x'))$ for the near-duplicate evaluation dataset in Figure \ref{fig:new-neardupes-rank}, for both the baseline and stabilized deep ranking feature. Stability training significantly increases the feature robustness, as the distribution of feature distances becomes more concentrated towards 0. For instance, for the original feature 76\% of near-duplicate image pairs has feature distance smaller than 0.1, whereas this is 86\% for the stabilized feature, i.e. the stabilized feature is significantly more similar for near-duplicate images.

\begin{figure*}[!ht]
\begin{center}
\begin{subfigure}[t]{0.5\textwidth}
        \centering
        \begin{tabular}{cc|}
        % jpeg
        \includegraphics[width=0.22\textwidth,height=0.22\textwidth]{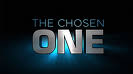} \includegraphics[width=0.22\textwidth,height=0.22\textwidth]{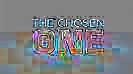}&
\includegraphics[width=0.22\textwidth,height=0.22\textwidth]{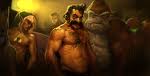} \includegraphics[width=0.22\textwidth,height=0.22\textwidth]{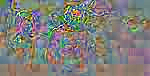}\\
 \Red{0.102} \ra\hspace{1pt} \Green{0.030} & \Red{0.107} \ra\hspace{1pt} \Green{0.048} \\
\includegraphics[width=0.22\textwidth,height=0.22\textwidth]{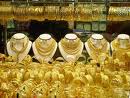} \includegraphics[width=0.22\textwidth,height=0.22\textwidth]{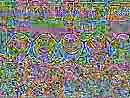}&
\includegraphics[width=0.22\textwidth,height=0.22\textwidth]{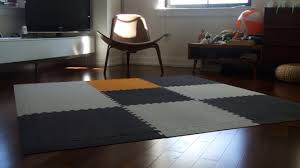} \includegraphics[width=0.22\textwidth,height=0.22\textwidth]{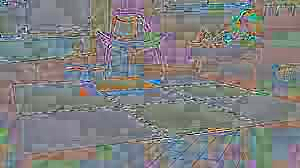}\\
\Red{0.105} \ra\hspace{1pt} \Green{0.039} & \Red{0.100} \ra\hspace{1pt} \Green{0.054} \\
\includegraphics[width=0.22\textwidth,height=0.22\textwidth]{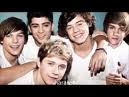} \includegraphics[width=0.22\textwidth,height=0.22\textwidth]{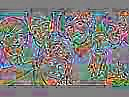}&
\includegraphics[width=0.22\textwidth,height=0.22\textwidth]{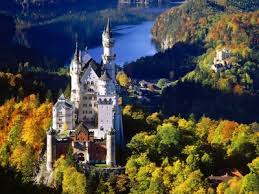} \includegraphics[width=0.22\textwidth,height=0.22\textwidth]{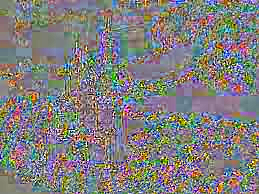}\\
\Red{0.106} \ra\hspace{1pt} \Green{0.013} & \Red{0.104} \ra\hspace{1pt} \Green{0.055}
        \end{tabular}
    \end{subfigure}%
    \begin{subfigure}[t]{0.5\textwidth}
        \centering
        % croppping
        \begin{tabular}{cc}
\includegraphics[width=0.22\textwidth,height=0.22\textwidth]{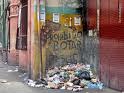} \includegraphics[width=0.22\textwidth,height=0.22\textwidth]{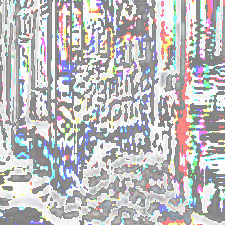}&
\includegraphics[width=0.22\textwidth,height=0.22\textwidth]{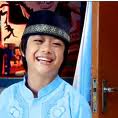} \includegraphics[width=0.22\textwidth,height=0.22\textwidth]{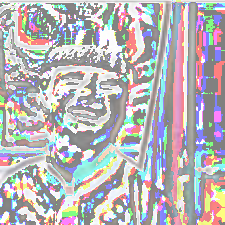}\\
\Red{0.128} \ra\hspace{1pt} \Green{0.041} & \Red{0.131} \ra\hspace{1pt} \Green{0.072} \\
\includegraphics[width=0.22\textwidth,height=0.22\textwidth]{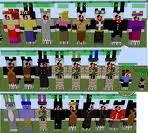} \includegraphics[width=0.22\textwidth,height=0.22\textwidth]{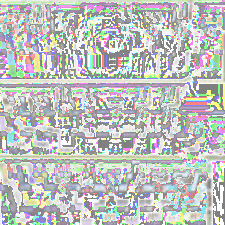}&
\includegraphics[width=0.22\textwidth,height=0.22\textwidth]{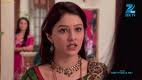} \includegraphics[width=0.22\textwidth,height=0.22\textwidth]{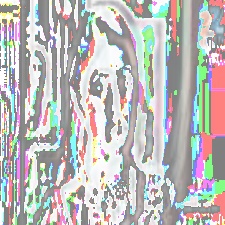}\\
\Red{0.122} \ra\hspace{1pt} \Green{0.068} & \Red{0.120} \ra\hspace{1pt} \Green{0.062} \\
\includegraphics[width=0.22\textwidth,height=0.22\textwidth]{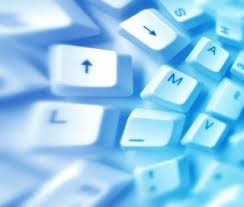} \includegraphics[width=0.22\textwidth,height=0.22\textwidth]{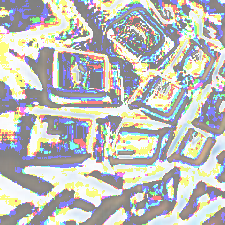}&
\includegraphics[width=0.22\textwidth,height=0.22\textwidth]{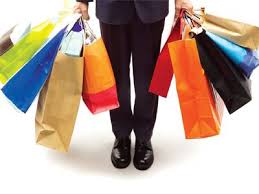} \includegraphics[width=0.22\textwidth,height=0.22\textwidth]{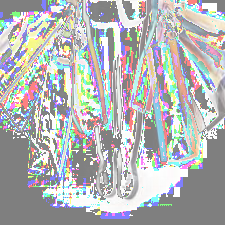}\\
\Red{0.150} \ra\hspace{1pt} \Green{0.079} & \Red{0.125} \ra\hspace{1pt} \Green{0.077}
        \end{tabular}
    \end{subfigure}
\end{center}
\vspace{-15pt}
\caption{Examples of near-duplicate image pairs that are robustly recognized as near-duplicates by stabilized features (small feature distance), but easily confuse un-stabilized features (large feature distance). Left group: using \jpeg{}-50 compression corruptions. Right group: random cropping \crop{}-10 corruptions. For each image pair, we display the reference image $x$, the difference with its corrupted copy $x-x'$, and the distance in feature space $D\brck{f(x), f(x')}$ for the un-stabilized (red) and stabilized features (green). %\corr{More in supplemental material?}
% \crop{}-10 was \crop{}-235
}
\label{fig:new-neardupes-class}
\end{figure*}

\tb{Stabilized feature distance.}
We also present our qualitative results to visualize the improvements of the stabilized features over the original features.
In Figure \ref{fig:new-neardupes-class} we show pairs of images and their \JPEG{} versions that were confusing for the un-stabilized features, i.e. that lay far apart in feature space, but whose stabilized features are significantly more close. This means that they are correctly detected as near-duplicates for much more aggressive, that is, lower detection thresholds by the stabilized feature, whereas the original feature easily confuses these as dissimilar images.
Consistent with the intuition that Gaussian noise applies a wide range of types of perturbations, we see improved performance for a wide range of perturbation types.
Importantly, this includes even localized, structured perturbations that do not resemble a typical Gaussian noise sample.

\subsection{Similar image ranking}
\label{sec:tripletranking}

The stabilized deep ranking features (see section \ref{sec:stab-feat-emb}) are evaluated on the similar image ranking task. Hand-labeled triplets from \cite{wang_learning_2014}\footnote{\url{https://sites.google.com/site/imagesimilaritydata/}.} are used as evaluation data. There are 14,000 such triplets. The ranking score-at-top-$K$ ($K=30$) is used as evaluation metric. The ranking score-at-top-$K$ is defined as
%
%\footnote{A subset of 5,000 triplet labels is available at \url{https://sites.google.com/site/imagesimilaritydata/}.}
%
%For our experiments with stable features, we used ranking features learned through the deep ranking network (see \ref{sec:network}). To this end, we obtained the triplet dataset that was used in \cite{wang_learning_2014}. The training dataset contains 14 million images with a learned pair-wise relevance measure, that acts as a proxy for the ground-truth similar-image ranking labels.
%Additionally, the triplet ranking dataset includes a set of 14,000 triplet ranking labels obtained from human raters and that was only used during evaluation. A subset of 5,000 triplet labels is publicly available on \url{https://sites.google.com/site/imagesimilaritydata/}.
%
%\tb{Evaluation metric.}
% We evaluated the ranking performance of the stabilized features $f_s$ obtained from stability training and unstable features $f$ by evaluating condition \refn{eq:feature-ranking} on the triplet evaluation dataset.
%
%For similar-image ranking, the ranking score @top-$K$ is defined as
%
\eq{\label{eq:ranking-score-def}
&\trm{ranking score @top-}K = \notag\\
&\trm{ \# correctly ranked triplets} - \trm{\# incorrectly ranked triplets},
}
where only triplets whose positive or negative image occurs among the closest $K$ results from the query image are considered. This metric measures the ranking performance on the $K$ most relevant results of the query image. We use this evaluation metric because it reflects better the performance of similarity models in practical image retrieval systems as users pay most of their attentions to the results on the first few pages.

\subsubsection{Experimental results.}
Our results for triplet ranking are displayed in Table \ref{table:ranking-score}. The results show that applying stability training improves the ranking score on both the original and transformed versions of the evaluation dataset.
The ranking performance of the baseline model degrades on all distorted versions of the original dataset, showing that it is not robust to the input distortions.
In contrast, the stabilized network achieves ranking scores that are higher than the ranking score of the baseline model on the \ti{original} dataset.

% \corrThis can be qualitatively understood by comparing the explicit robustness objective of stability training and the triplet ranking loss. For state-of-the-art similar-image ranking performance, the deep ranking network features have to correctly rank \ti{hard} triplets, where either a) the query-negative pair $(q,p)$ and query-negative pair $(q,n)$ are both visually similar or b) the positive-negative pair $(p.n)$ are near-duplicates. Stability training \ti{explicitly} implements the objective that similar images have similar features, that, it implements $D(f(x), f(x'))\approx 0$ for near-duplicates $x,x'$. In contrast, the triplet loss \refn{eq:hingeloss} implements a weaker condition by only requiring the feature distance $D(f(q),f(p))$ to be smaller than the margin $g$ than any other distance $D(f(q), f(n))$. Stability training can therefore yield more robust features that more easily rank hard triplets correctly. }

\begin{table}[!t]
\begin{center}
\begin{tabular}{p{55pt}|c|c}
Distortion & Deep ranking & Deep ranking + ST \\
\hline
Original & 7,312 & \tb{7,368} \\
\textsc{jpeg}-50 & 7,286 & \tb{7,360} \\
\textsc{thumb}-30k  & 7,160 & \tb{7,172} \\ % was \textsc{thumb}-180
\textsc{crop}-10 & 7,298 & \tb{7,322}
%crop at $235\times 235$ to 225\times 225
% thumbnail resize
\end{tabular}
% Transpose
% \begin{tabular}{l|c|c|c|c}
% Model & Original & JPEG 50\% & Thumbnail $180\times 180$ & Random cropping $235\times 235$ \\
% %crop at $235\times 235$ to 225\times 225
% % thumbnail resize
% \hline
% Deep ranking & 7312 & 7286 & 7160 & 7298\\
% Deep ranking with ST  & \tb{7368} & \tb{7360} & \tb{7172} & \tb{7322}
% \end{tabular}
% Data augmentation with Gaussian noise & 5514 & & & \\
% Stability training: $\alpha=0.1, \sigma=0.1$ & \tb{7368} & \tb{7360} & 7300  \\
% Stability training: $\alpha=0.1, \sigma=0.2$ & 7346 & 7294 & \tb{7322}
\caption{Ranking score @top-30 for the deep ranking network with and without stability training (higher is better) on distorted image data. Stability training increases ranking performance over the baseline on all versions of the evaluation dataset. We do not report precision scores, as in \cite{wang_learning_2014}, as the ranking score @top-30 agrees more with human perception of practical similar image retrieval.}
\label{table:ranking-score}
\end{center}
\end{table}

\subsection{Image classification}

In the classification setting, we validated stability training on the ImageNet classification task \cite{ILSVRC15}, using the Inception network \cite{szegedy_going_2015}. We used the full classification dataset, which covers 1,000 classes and contains 1.2 million images, where 50,000 are used for validation.
We evaluated the classification precision on both the original and a \JPEG-50 version of the validation set. Our benchmark results are in Table \ref{tab:imagenet-p-at-5}.

Applying stability training to the Inception network makes the class predictions of the network more robust to input distortions. On the original dataset, both the baseline and stabilized network achieve state-of-the-art performance. However, the stabilized model achieves higher precision on the distorted evaluation datasets, as the performance degrades more significantly for the baseline model than for the stabilized model. For high distortion levels, this gap grows to 5\% to 6\% in top-1 and top-5 precision.

\begin{table}[!]
\begin{center}
\begin{tabular}{l|c|c|c}
Precision @top-5 & Original & \JPEG-50 & \JPEG{}-10 \\  % & Random cropping $235\times 235$ & Thumbnail resize \\
\hline
Szegedy et al \cite{szegedy_going_2015} & 93.3\% & & \\
\hline
Inception & \tb{93.9\%}  & 92.4\% & 83.0\% \\
Stability training & 93.6\%  & \tb{92.7\%} & \tb{88.3\%} \\
&&&\\
Precision @top-1 & && \\  % & Random
\hline
Inception & 77.8\% & 75.1\% & 61.1\%\\
Stability training & \tb{77.9\%} & \tb{75.7\%} & \tb{67.9\%} %
\end{tabular}

\caption{Classification evaluation performance of Inception with stability training, evaluated on the original and \jpeg{} versions of ImageNet. Both networks give similar state-of-the-art performance on the original evaluation dataset (note that the performance difference on the original dataset is within the statistical error of $0.3\%$ \cite{ILSVRC15}).
However, the stabilized network is significantly more robust and outperforms the baseline on the distorted data.
}
\label{tab:imagenet-p-at-5}
\end{center}
\end{table}

\begin{figure}[!t]
\centering
\includegraphics[width=0.48\textwidth]{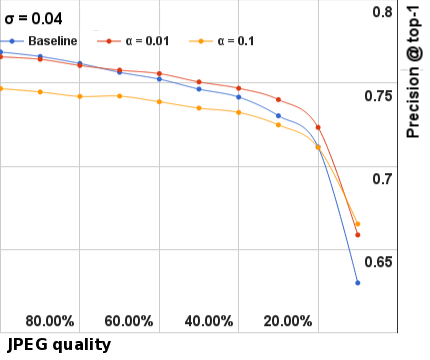}
\caption{A comparison of the precision @ top-1 performance on the ImageNet classification task for different stability training hyper-parameters $\alpha$, using \JPEG{} compressed versions of the evaluation dataset at decreasing quality levels, using a fixed $\sigma=0.04$. At the highest \JPEG{} quality level, the baseline and stabilized models perform comparably. However, as the quality level decreases, the stabilized model starts to significantly outperform the baseline model.}
% This qualitative behavior is visible for a wide range of hyper-parameters $\alpha, \sigma$: for instance, using $\alpha = 0.01$ and $\sigma = 0.04$ results in better performance already below the 80\% quality level. }
\label{fig:performance-response}
\end{figure}

\tb{Robust classification on noisy data.}
We also evaluated the effectiveness of stability training on the classification performance of Inception on the ImageNet evaluation dataset with increasing \JPEG{} corruption. In this experiment, we collected the precision @top-1 scores at convergence for a range of the training hyper-parameters:
the regularization coefficient $\alpha$ and noise standard deviation $\sigma$. A summary of these results is displayed in Figure \ref{fig:performance-response}.

At the highest \JPEG{} quality level, the performance of the baseline and stabilized models are comparable, as the visual distortions are small. However, as the \JPEG{} distortions become stronger, the stabilized model starts to significantly outperform the baseline model. This qualitative behavior is visible for a wide range of hyper-parameters, for instance, using $\alpha = 0.01$ and $\sigma = 0.04$ results in better performance already below the 80\% quality level.
 \tempnewpage
\section{Conclusion}\label{sec:conclusion}

In this paper we proposed stability training as a lightweight and effective method to stabilize deep neural networks against natural distortions in the visual input. Stability training makes the output of a neural network more robust by training a model to be constant on images that are copies of the input image with small perturbations.
As such, our method can enable higher performance on noisy visual data than a network without stability training.
We demonstrated this by showing that our method makes neural networks more robust against common types of distortions coming from random cropping, JPEG compression and thumbnail resizing. Additionally, we showed that using our method, the performance of stabilized models is significantly more robust for near-duplicate detection, similar-image ranking and classification on noisy datasets.

% \subsection{Discussion / Future work}

% \iitem{
% \item Nonlinear models
% \item Relation to adversarial training: adversarial examples are structured noise. Do they infer stability? Use noisy samples where
% \eq{\brcka{\sigma_k, \fr{dJ}{dx_k}}=1}
% \item Theory?
% }

% \ti{Acknowledgements}
 \tempnewpage

{\small
\bibliographystyle{ieee}
\bibliography{paper}
}

\end{document}